\documentclass[10pt]{cai26}

\usepackage{tabularx}
\usepackage{xcolor}
\usepackage{graphicx}
\usepackage{subcaption}
\usepackage{booktabs}
\usepackage{array}
\usepackage{siunitx}
\usepackage{colortbl}
\usepackage{enumitem}

\begin{document}

\newcommand{\yifan}[1]{{\color{black}{{\normalsize   #1}}}}
\newcommand{\mc}[1]{\textcolor{black}{#1}}
\newcommand{\jack}[1]{\textcolor{black}{#1}}
\newcommand{\CameraReady}[1]{\textcolor{black}{#1}}

\def\conferenceyear{2026}
\volumeheader{39}{0}
\begin{center}

\title{NameBERT: Scaling Name-Based Nationality Classification with LLM-Augmented Open Academic Data}
\maketitle

\thispagestyle{empty}
\pagenumbering{gobble}

\begin{tabular}{c}
Cong Ming\upstairs{\affilone}, 
Ruixin Shi\upstairs{\affilone}, 
Yifan Hu\upstairs{\affilone,*}
\\[0.25ex]
{\small \upstairs{\affilone} Northeastern University, United States}
\end{tabular}

\emails{
  \upstairs{*} yifanh@gmail.com
}
\vspace*{0.1in}
\end{center}

\begin{abstract}
Inferring nationality from personal names is a critical capability for equity and bias monitoring, personalization, and a valuable tool in biomedical and sociological research. However, existing name-based nationality classifiers are 
    \mc{typically trained on relatively small or source-specific labeled datasets, which can introduce coverage gaps and limit performance for underrepresented countries.}
    While large language models (LLMs) demonstrate strong zero-shot performance for name-based nationality prediction, their computational cost and latency make them impractical for real-time, large-scale deployment. In this work, we created a large-scale name-nationality dataset from the Open Academic Graph (OAG) and introduce a framework that leverages LLMs as dataset enrichers rather than inference engines. 
    \mc{We augment low-resource countries with LLM-generated names and evaluate on real and synthetic-tail test sets. We find that augmentation produces large gains when evaluation includes synthetic tail names and still offers a modest lift on tail-country metrics otherwise. Overall, NameBERT models achieve significantly higher accuracy than state-of-the-art baselines across both in- and out-of-domain tasks, while remaining efficient for large-scale inference compared to LLMs.} 
\end{abstract}

\copyrightnotice


\section{Introduction}
    Names carry rich cultural and demographic signals, and inferring such information is central to a wide range of applications, from equity and bias monitoring to improving the quality and personalization of online services. As the direct collection of sensitive demographic information becomes increasingly constrained by privacy regulations, inferring demographics from names has emerged as a practical and widely used proxy when individual-level data are unavailable. In this work, we focus on inferring nationality from names.
    
    A substantial body of work has studied name-based nationality inference. For example, NamePrism~\cite{ye2017nationality}, which was trained on approximately 74 million entries using a combination of name embeddings and Naive Bayes techniques, achieved strong performance in nationality prediction. However, it used proprietary data, making the study difficult to replicate. Subsequent methodologies have transitioned toward modeling names as character sequences. Early sequence-based approaches such as Name2Nat~\cite{park2020name2nat,sood2018predicting} used bidirectional GRUs or LSTM. Building upon these, transformer-based methods like RaceBERT~\cite{parasurama2021racebert} have further improved accuracy by fine-tuning pretrained models, while frameworks such as EthnicIA~\cite{jain2022importance} emphasize the interpretability of demographic inferences. However, existing literature faces limitations regarding data scope and granularity. Most of these models focus on ethnicity prediction, and rely on regional or selective training data, limiting their capacity to infer fine-grained nationalities. On the other hand, while models such as NamePrism~\cite{ye2017nationality}, Name2Nat~\cite{park2020name2nat}, and EthnicSeer~\cite{treeratpituk2012name} aim to predict finer-grained national or ethnic categories, they are often constrained by a limited number of output classes (e.g., 12 classes for EthnicSeer) or exhibit poor performance on long-tail countries due to insufficient training data.

Our work is motivated by three key considerations. First, we believe that a larger, higher-quality labeled dataset would greatly benefit model training. Second, we assess whether models trained on this data can generalize beyond the training distribution, performing well on both in-domain and out-of-domain evaluations. Third, to address the label imbalance that inevitably appears in most datasets, we investigate whether LLMs can serve as dataset enrichers by generating additional supervision for sparsely represented countries, rather than acting as inference engines.
In this work, we construct a dataset of 1.34M names with proxy nationality labels derived from authors’ affiliation data in the Open Academic Graph (OAG)~\cite{tang2008arnetminer, sinha2015overview}. 
\mc{We further leverage LLMs to generate names for low-resource countries and study the conditions under which such augmentation is beneficial.}

We study three research questions: 
\textbf{RQ1}: Can OAG be leveraged to provide a large and diverse source of labels for nationality classification? \textbf{RQ2}: 
\mc{How do neural models trained on OAG compare to state-of-the-art nationality classifiers across in-domain and out-of-domain evaluations? And
\textbf{RQ3}: Does LLM-augmented OAG data help to improve model performance?}
Our contributions are as follows:
\begin{itemize}[leftmargin=*, nosep]
    \item We curate a new dataset based on OAG containing over 1.34M names with proxy nationality labels derived from author affiliation metadata; we will open-source this dataset and the models upon publication of the paper.
    \item We propose an LLM-based enrichment pipeline 
    \mc{for low-resource countries and provide an empirical analysis of its distribution-dependent effects across evaluation settings.}
    \item 
    \mc{We develop transformer-based classifiers, NameBERT, and benchmark them against state-of-the-art models (e.g., NamePrism~\cite{ye2017nationality}, EthnicSeer~\cite{treeratpituk2012name}, and Name2Nat~\cite{park2020name2nat}). We demonstrate superior performance on real-name and out-of-domain benchmarks. We also clarify the settings in which LLM-based augmentation yields gains.}
\end{itemize}

\section{Related Work}
\textbf{Ethnicity Models.} Early name-based demographic inference models focused primarily on ethnicity prediction rather than nationality. Dominant sequence-based and transformer-based approaches, such as those proposed by Sood et al.~\cite{sood2018predicting} and Parasurama et al.~\cite{parasurama2021racebert}, model names as character sequences using machine learning methods. 

These methods improved robustness to spelling variations and name morphology. However, 
\mc{many widely used ethnicity benchmarks are U.S.-centric (e.g., derived from voter registrations), which limits geographic coverage} and 
\mc{often targets race/ethnicity categories rather than national origin.}
As a result, models trained on this distribution exhibit substantial geographic and representational bias and are unsuitable for fine-grained, global nationality prediction.

\textbf{Fine-Grained Nationality Models.} NamePrism~\cite{ye2017nationality} was trained on $\sim$74M names from proprietary contact lists and Twitter. Instead of predicting countries, it maps names into a fixed 39-leaf taxonomy of broader regional/ethnic groups, which limits fine-grained national origin inference, especially for underrepresented regions. Its reliance on proprietary data also limits reproducibility. Other global approaches, such as Name2Nat~\cite{park2020name2nat} and EthnicSeer~\cite{treeratpituk2012name}, attempt to predict nationalities using Wikipedia-derived datasets. Name2Nat is trained on the NaNa dataset, 
constructed by parsing the English Wikipedia dump to extract names spanning 173 countries. However, this dataset suffers from extreme class imbalance: many nationalities have limited samples (e.g., ``Laos'' has 26 training, 3 validation, and 4 test instances), leading to poor generalization for long-tail countries.


\textbf{LLMs for nationality prediction.} Recently, LLMs have demonstrated strong zero-shot performance on demographic inference~\cite{alnuaimi2024whats, shang2025fairness}. Their semantic reasoning capabilities allow them to rival or surpass supervised baselines in certain settings. For example, AlNuaimi et al.~\cite{alnuaimi2024whats} explored zero-shot demographic prediction using LLMs, reporting strong results for gender prediction and competitive performance for ethnicity classification. 
However, direct deployment of LLMs as classifiers is computationally expensive and impractical for large-scale or real-time applications involving millions of records. 

\mc{Overall, prior work either relies on geographically narrow or proprietary data, restricts output granularity, or uses LLMs directly as inference engines, leaving open the question of how to build scalable, fine-grained nationality classifiers with broad coverage.}

\section{Dataset}
The Open Academic Graph (OAG)~\cite{tang2008arnetminer,sinha2015overview} is a large-scale academic knowledge graph that integrates data from multiple sources and provides comprehensive metadata for over 130 million publications and 35 million authors. The author affiliation information in OAG serves as strong signals for an author's nationality or geographic origin, particularly when an author has a consistent publication history tied to institutions in specific countries. The academic domain also offers several advantages: (1) author names in OAG are typically more standardized and complete than social media usernames or informal text; (2) the dataset spans diverse linguistic and cultural backgrounds, providing broad cross-country coverage. 

Since OAG does not explicitly include nationality annotations, we design a heuristic label extraction pipeline that derives nationality labels from institutional affiliation information. We describe this pipeline in detail and characterize the resulting labeled dataset. We acknowledge that affiliation does not always coincide with nationality. For example, researchers may work abroad, change institutions over time, or hold multiple nationalities. However, given the size, diversity, and longitudinal nature of OAG, these sources of noise are mitigated in aggregate (Section~\ref{sec:data_charactrirtics}). As a result, OAG remains a valuable foundation for constructing large-scale training data for robust name-to-nationality classification.

\subsection{Extracting nationality labels from OAG}
    
We observe that affiliation strings often end with a country name after a comma. 
We therefore apply a regular-expression rule that captures the trailing token after the final comma and treats it as a candidate country label. This rule correctly extracts country labels from affiliation strings such as ``University of Oxford, {UK}''. Extracted tokens are further validated against abbreviations (e.g., ``USA'', etc.) and native endonyms (e.g., ``Deutschland'', ``Nippon'', etc.). This normalization process yielded approximately 1.59 million raw name--country pairs. If an author lists multiple affiliations associated with different countries, we exclude that entry to avoid ambiguity in country assignment. After removing ambiguous entries, we retained a final dataset of 1.34M unique, high-confidence labeled pairs spanning 99 countries.

OAG country labels are proxy annotations derived from affiliation metadata and could be noisy. To estimate plausibility, we run an LLM-based validation on sampled name-country pairs, asking GPT-4o to judge whether the name is reasonably common for the claimed country. For countries with distinctive naming conventions, we apply a stricter acceptance rule: the name must be clearly characteristic of or strongly associated with that country, rather than merely plausible. For multicultural countries, we instead apply a plausibility criterion, allowing names that could reasonably occur in the country’s population even if not majority-typical. Across 99 countries, GPT-4o accepts 84.14\% of pairs (ranging from $\approx$ 95\% for East Asian countries to $\approx$ 60\% for African countries), suggesting that most of the labels are plausible, while leaving some residual noise; therefore, we present results on both the raw split and a filtered evaluation split with LLM-screened labels.





\subsection{Representativeness Analysis of Name Distributions}
\jack{
We compare first-name frequencies in OAG to U.S.\ Social Security Administration baby name records (1954--1999). For the 100 most common Census names, we compute the ratio of each name’s frequency in OAG\_US (U.S.-affiliated authors) to its Census frequency. 
}


Across the 100 most common Census names, the median ratio is 0.54, below parity because 12.4\% of OAG\_US entries use abbreviated given names (e.g., \textit{Q.}) and 10.3\% use names absent from SSA records (e.g., \textit{McCarthy}, \textit{Yin}, \textit{Tsai}). Despite this, the rank ordering of common Census names is largely preserved in OAG\_US. Several older names - \textit{Paul} (1.01), \textit{Stephen} (1.04), \textit{George} (1.08) - reach or exceed parity, while newer cohort names such as \textit{Ashley} (0.23), \textit{Brittany} (0.12), and \textit{Brandon} (0.20) fall well below, reflecting the lag from academic entry to publications.


Since the dataset draws exclusively from academic publications, it skews toward formal naming conventions typical of highly educated populations. Diminutive variants, e.g., \textit{Dick}, \textit{Bob} are indeed exceedingly rare in OAG, whereas formal counterparts \textit{Richard} (0.69), \textit{Robert} (0.67) are more common, reflecting academic naming conventions. However, this underrepresentation reflects a domain-specific convention of academic publishing rather than a socioeconomic signal - academics who go by \textit{Dick} in everyday life still publish as \textit{Richard}. Overall, disparities are driven primarily by metadata structure and cohort composition rather than social stratification.


\subsection{Dataset Characteristics\label{sec:data_charactrirtics}}

\jack{
Despite the large volume of extracted pairs, the nationality distribution in OAG is highly long-tailed (Figure~\ref{fig:data_distribution}). A small number of countries (e.g., China, USA, France) dominate the dataset, while 10 countries (e.g., Bhutan, North Korea) have fewer than 50 verified entries. Such imbalance limits standard supervised classifiers, motivating our LLM-driven data augmentation framework.
}

\jack{
We also examined name duplication across countries. Approximately 17,000 names ($\sim$1.2\%) appear in at least two countries, and about 1,600 ($\sim$0.1\%) appear in three or more. Among duplicated names, Chinese names account for roughly 45\%, followed by French (6.4\%) and German (5.6\%).
Some duplication reflects cross-border academic mobility rather than true name ambiguity. In our pipeline, authors linked to multiple countries are removed; however, researchers who publish exclusively from a single foreign institution remain labeled by affiliation country. Given the small proportion of duplicated names, this residual noise is limited. In practice, NameBERT tends to predict the statistically dominant country. For example, ``Y.\ Wang'' appears in 14 countries (56.5\% China), and the model predicts ``china'' with confidence 0.553, aligning with the majority distribution. This behavior is appropriate for population-level analysis.
}





\section{Methodology\label{sec_metholodogy}}

Building on the class imbalance identified in Section~3, we introduce a hybrid framework that leverages large language models (LLMs) as {data enrichers} rather than end-to-end classifiers. 
\mc{Specifically, we use LLMs to generate additional names for low-resource countries
to increase training coverage in the long tail. Importantly, we do not assume such augmentation universally improves generalization; instead, we evaluate its impact across in-domain real-name benchmarks, synthetic-mixed tail diagnostics, and out-of-domain test sets. 
}


\subsection{\mc{Zero-Shot Name-Only Reference Performance}}

\mc{To contextualize ML model performance, we evaluate three LLMs in a zero-shot setting where nationality is predicted directly from full names, restricted to the same 99-country label space. Evaluation is conducted on a balanced mini test set (up to 50 names per country sampled from test\_oag). The balanced design prevents large countries from dominating the metric and isolates cross-country name signal strength.}

\mc{We emphasize that this setting does not represent a strict performance ceiling. Instead, it serves as a distribution-agnostic reference capturing how much nationality information is recoverable from names alone without dataset-specific training.}
\yifan{Predicting nationality across 99 fine-grained countries is inherently challenging, as many full names are shared among neighboring or culturally related nations (e.g., ``Andrey Ivanov'' appears in both Ukraine and Russia). Consistent with this ambiguity, when the 99 labels are aggregated into 39 and 12 broader groups, GPT-5's accuracy improves from 58.6\% to 75.3\% and 79.1\%, respectively. This pattern suggests that the lower accuracy in the 99-class setting mainly reflects confusion among closely related countries, rather than deficiencies in label quality.
}

\begin{table*}[htbp]
\vspace{-0.2cm}
\footnotesize
\centering
\sisetup{table-format=1.3}
\begin{tabular*}{\textwidth}{@{\extracolsep{\fill}} l c S S S}
\toprule
\textbf{Model} & \textbf{Class} & {\textbf{Accuracy}} & {\textbf{Weighted-F1}} & {\textbf{Macro-F1}} \\
\midrule
LLM (GPT-4o)        & 99 & 0.544 & 0.539 & 0.488 \\
LLM (Claude-3.5)   & 99 & 0.465 & 0.501 & 0.454 \\
\textbf{LLM (GPT-5)} & 99 & \textbf{0.586} & \textbf{0.574} & \textbf{0.521} \\
\arrayrulecolor{gray}
\midrule
\arrayrulecolor{black}  
LLM (GPT-5) & 39 & 0.753 & 0.756 & 0.757 \\
\arrayrulecolor{gray}
\midrule
\arrayrulecolor{black}  
LLM (GPT-5) & 12 & 0.791 & 0.817 & 0.760 \\
\bottomrule
\end{tabular*}
\vspace{-0.3cm}
\caption{Performance of LLMs on 4K \textbf{test\_oag} samples at different label granularities.}
\label{tab:results-random-sampled}
\vspace{-0.6cm}
\end{table*}




\subsection{LLM-based Data Augmentation}
We use LLMs to augment training data for underrepresented nationalities, instead of directly at inference time. The objective of this augmentation is not to \CameraReady{equalize all classes to a common size, but to provide additional supervision for underrepresented countries.}

For each nationality with fewer than 6{,}000 instances, we prompt {GPT-5} to generate \CameraReady{up to} 5{,}000  
additional plausible full names to enrich the long tail (\CameraReady{approximately} 3{,}000/1{,}000/1{,}000 train, validation and test split; \CameraReady{see Figure~\ref{fig:split} for exact sizes}). 
The template is structured to encourage name diversity and formatting consistency: 
\textit{``Generate \texttt{\{N\}} realistic full names for people from \texttt{\{country\}}. 
Each line should contain a unique full name (first \& last name). 
Avoid repeating the same first or last names more than 3 times.''}
Importantly, augmentation is applied exclusively to classes below the 6{,}000-sample threshold; majority classes are left unchanged to avoid distorting the overall data distribution. \emph{To prevent data leakage, we ensure that no names from the validation or test splits appear in the training data.}

\subsection{Training and Evaluation Protocol}



\begin{figure}[t]
\vspace{-0.5cm}
  \centering
  \includegraphics[width=\textwidth]{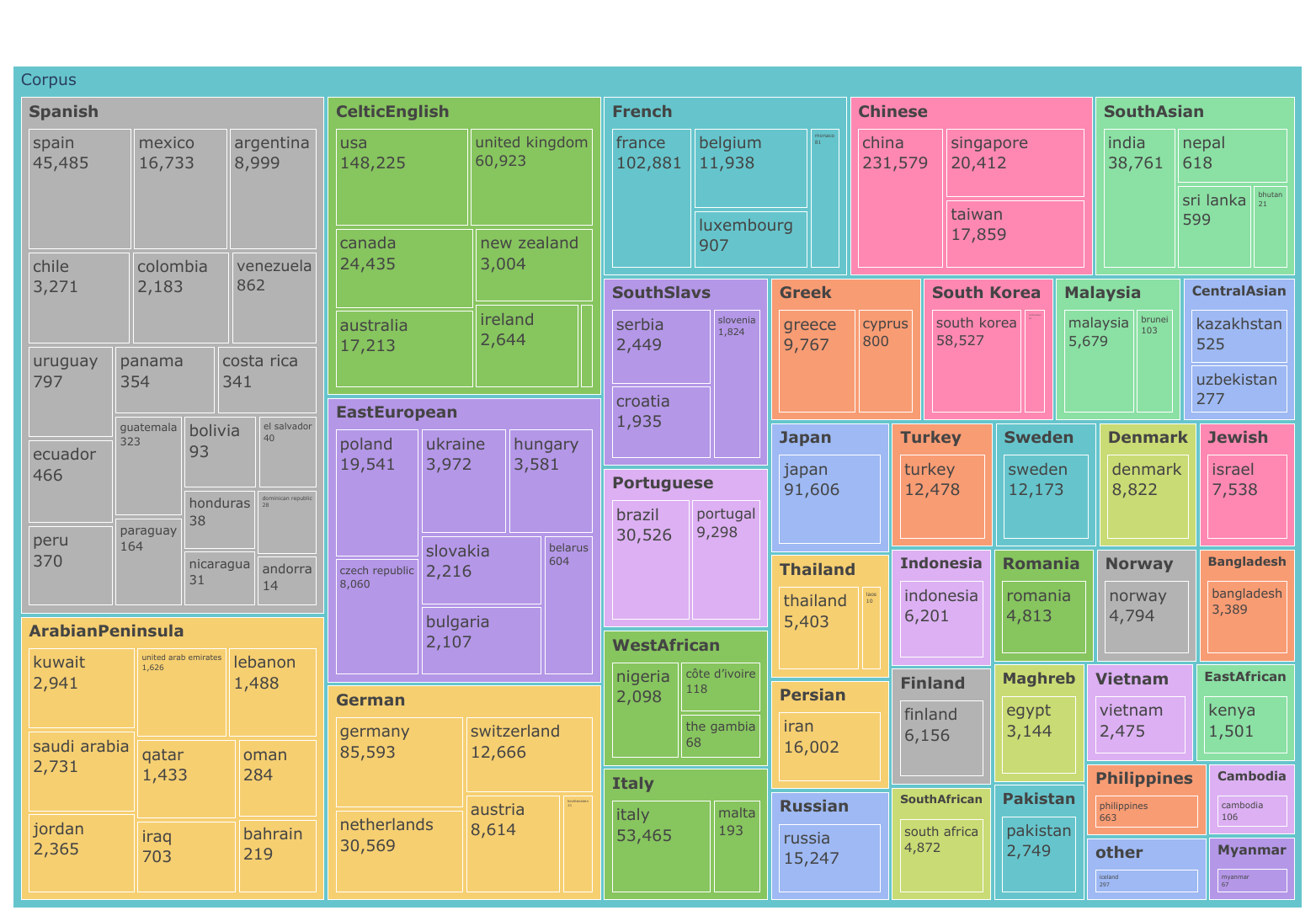}
  \vspace{-0.7cm} 
  \caption{1.34M OAG dataset of 99 classes mapped to NamePrism taxonomy of 39 classes}
  \label{fig:data_distribution}
  \vspace{-0.5cm} 
\end{figure}

To compare models with different label spaces, we apply directional label mappings only at evaluation time: Name2Nat’s 173 labels are mapped into our 99-class space, while our predictions are mapped into the coarser taxonomies used by EthnicSeer (12 classes) and NamePrism (39 classes). Figure~\ref{fig:data_distribution} illustrates the mapping from our 99 classes to NamePrism’s 39-class taxonomy. These mappings are used solely for evaluation and do not affect training, enabling fair comparison across heterogeneous output granularities.

\subsection{Model Training}



Our model, \textbf{NameBERT}, fine-tunes \texttt{bert-base-multilingual-cased}. Names are tokenized using WordPiece, and the final-layer [CLS] representation is passed to a linear classification head over $C$ nationalities. We cap the maximum sequence length at 40 subword tokens, which covers most names while minimizing computational overhead. 
We denote the model trained on the original OAG data as NameBERT$_{oag}$, and the model trained on LLM-augmented OAG data as NameBERT$_{aug}$.

    \subsection{Baselines \label{baselines}}

We compare our model with three state-of-the-art classifiers. \newline \indent \textbf{EthnicSeer}~\cite{treeratpituk2012name} formulates name--ethnicity classification 
    as a supervised learning problem. It uses a logistic regression classifier trained on approximately 215K Wikipedia names grouped into 12 ethnicity categories, and reports an accuracy of about 0.85 in the original paper. We use its Python library~\cite{ethnicseer}.

    \textbf{NamePrism}~\cite{ye2017nationality} expands the label space to 39 nationality groups and employs a {Na\"{i}ve Bayes} classifier trained on large-scale email contacts and Twitter data (approximately 68M + 6M names). The authors report a weighted F1 score of 0.806 on their test data, and NamePrism remains a strong benchmark for large-coverage nationality prediction. We queried the model API~\cite{nameprism}.
    
    \textbf{Name2Nat}~\cite{park2020name2nat} is a Python package for nationality prediction based on the NaNa dataset (approximately 1.1M Wikipedia names covering 170+ countries). It uses a bidirectional GRU architecture. The model predicts one of 170+ country labels from Romanized names and serves as a modern neural baseline.

\section{Experiment}

In this section, we describe our dataset validation and augmentation pipeline, experimental setup, and evaluation methodology designed to assess both overall performance and long-tail robustness. \yifan{Our goal is to evaluate whether NameBERT trained on OAG data performs well across in-domain and out-of-domain settings, and whether/when LLM augmentation to address data imbalance further boosts performance.}

\subsection{Data Splits, Label Validation and Data Augmentation}

\begin{figure}[htbp]
\vspace{-0.6cm}
      \centering
      \includegraphics[width=0.85 \textwidth]{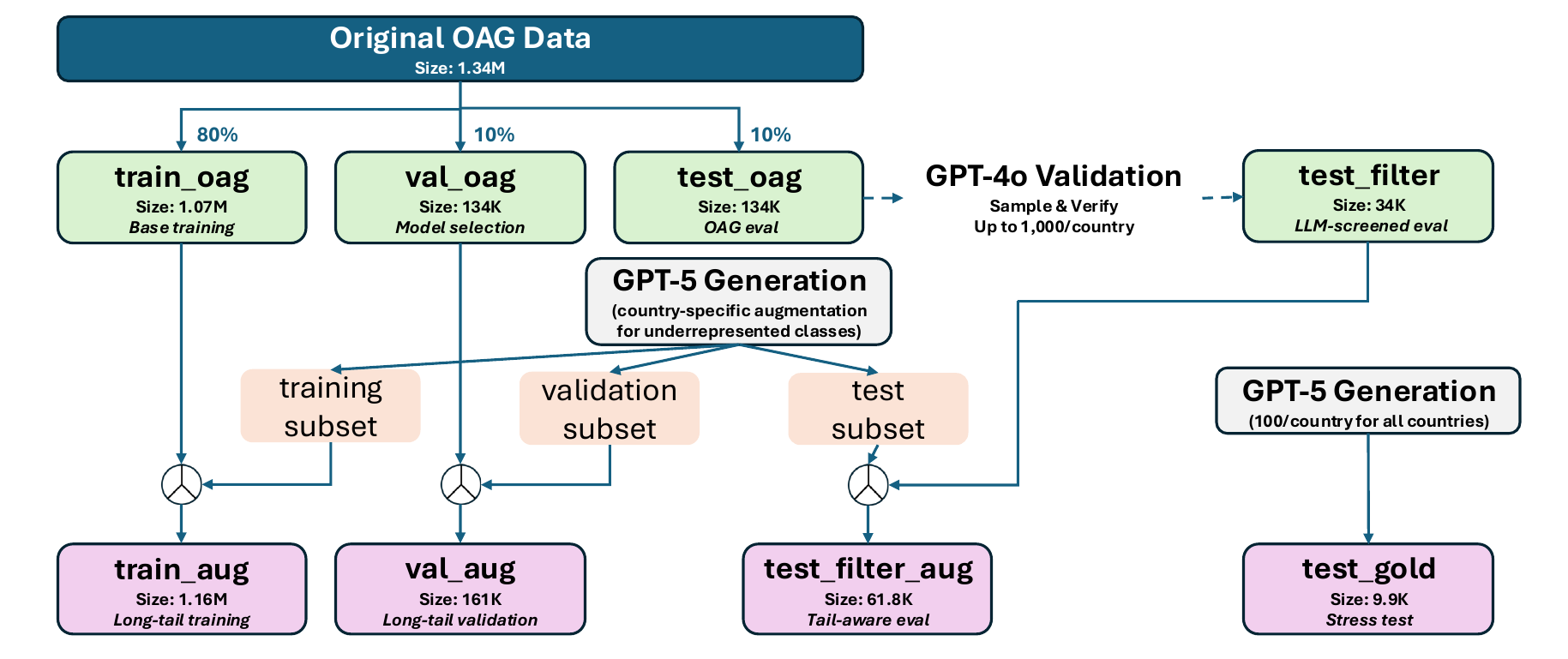}
      \vspace{-0.2cm}
      \caption{Dataset construction pipeline, splits, sizes, and purposes. Synthetic augmentation uses country-specific generation budgets.}
      \label{fig:split}
      \vspace{-0.5cm}
\end{figure}

We train and evaluate models on multiple dataset splits derived from the 1.34M OAG corpus, and from the LLM validated and augmented versions of this corpus. Figure~\ref{fig:split} summarizes the construction pipeline, all splits and their roles.
 We emphasize that \emph{all test data, including synthetic test names, are excluded from training to prevent label leakage.}

\textbf{Original OAG data.} Our starting dataset is the 1.34M OAG name-nationality corpus, which is split 8/1/1 into \texttt{train\_oag}, \texttt{val\_oag}, and \texttt{test\_oag}. 

\textbf{Validated high-quality test data.} Because \texttt{test\_oag} preserves real-world label noise, we construct a higher-quality evaluation set, \texttt{test\_filter}, by sampling from \texttt{test\_oag} and using GPT-4o to verify whether each name is reasonably common for its assigned country. Sampling continues until either 1{,}000 validated names per country are obtained or all candidates are exhausted.

\textbf{LLM-augmented data.} 
To increase supervision for the long tail, we generate synthetic
names for 57 underrepresented countries using GPT-5 ($\approx$ 5{,}000 names per country, \CameraReady{subject to country-specific generation budgets}) and split them into train/val/test \CameraReady{following a 3:1:1 ratio}. We then merge these partitions into the corresponding splits: 
the training partition is added to \texttt{train\_oag} to form \texttt{train\_aug}, the validation partition to \texttt{val\_oag} to form \texttt{val\_aug} and the test partition to the LLM-screened test set \texttt{test\_filter} to form \texttt{test\_filter\_aug}.


\textbf{LLM synthetic test data.} 
\mc{In addition, we construct \texttt{test\_gold}, a fully synthetic
stress-test set comprising 100 names per country. This split is intended as a controlled diagnostic for probing cross-country name separability and sensitivity to tail classes, as well as a source of out-of-domain test data for all models except NameBERT$_{aug}$}. 

\subsection{Models and Training Setup}

NameBERT models are trained as multi-class classifiers minimizing cross-entropy loss. We consider two training settings throughout the paper: \textit{oag} and \textit{aug}, corresponding to training on \texttt{train\_oag} and \texttt{train\_aug}, respectively. Model selection is performed on the matched validation split (\texttt{val\_oag} or \texttt{val\_aug}).
The NameBERT models are fine-tuned using AdamW (learning rate $2\times10^{-5}$, batch size 64) for up to 10 epochs, with linear warmup over 10\% of training steps. 
The checkpoint with the best validation Macro F1 is selected. We report Accuracy, Weighted F1, and Macro F1. To ensure long-tail robustness, Macro F1 is used for early stopping (patience = 5) and emphasized in result comparisons.

\subsection{Results and Analysis}
We conduct experiments with three baseline models in Section~\ref{baselines}, and two supervised neural models: NameBERT$_{oag}$ and NameBERT$_{aug}$.



\begin{table}[htbp]
\vspace{-0.4cm}
\footnotesize
\centering
\begin{tabular}{l c c c c}
\toprule
\textbf{Model} & \textbf{Classes} & \textbf{Accuracy} & \textbf{Weighted-F1} & \textbf{Macro-F1} \\
\midrule
EthnicSeer & 12 & 0.677 & 0.670 & 0.548 \\
{NameBERT$_{oag}$} & 12 & \textbf{0.808} & \textbf{0.820} & 0.741 \\
{NameBERT$_{aug}$} & 12 & 0.806 & 0.819 & \textbf{0.742} \\
\midrule
NamePrism & 39 & 0.589 & 0.603 & 0.383 \\
{NameBERT$_{oag}$} & 39 & 0.787 & 0.785 & 0.604 \\
{NameBERT$_{aug}$} & 39 & \textbf{0.788} & \textbf{0.785} & \textbf{0.607} \\
\midrule
Name2Nat  & 99 & 0.467 & 0.442 & 0.139 \\
NameBERT$_{oag}$ & 99 & \textbf{0.699} & 0.676 & 0.349 \\
{NameBERT$_{aug}$} & 99 & 0.698 & \textbf{0.680} & \textbf{0.364} \\
\bottomrule
\end{tabular}
\vspace{-0.2cm}
\caption{Performance comparison on \texttt{test\_oag}. Models are compared at 12 classes (EthnicSeer taxonomy), 39 classes (NamePrism taxonomy), and 99 classes (full OAG taxonomy) using matched label mappings.}
\vspace{-0.4cm}
\label{tab:combined_results_oag}
\end{table}
\vspace{-0.4cm}

\subsubsection{Performance on OAG Data}

We first evaluate all models on \texttt{test\_oag}, a held-out split drawn from the 1.34M OAG dataset that preserves 
\mc{proxy-label noise from affiliation-derived country annotations. This setting serves as an in-domain evaluation scenario, where name--country associations may be imbalanced, noisy, or weakly supported.}

{Table~\ref{tab:combined_results_oag}} reports results under the full {99-class OAG taxonomy}. 
\mc{Both NameBERT models substantially outperform Name2Nat. Between the NameBERT variants, training with augmentation yields higher Weighted-F1 (0.680 vs. 0.676) and Macro-F1 (0.364 vs. 0.349), while Accuracy remains essentially unchanged (0.698 vs. 0.699). } 




\mc{To contextualize these results against two other baselines with reduced taxonomies, Table~\ref{tab:combined_results_oag} also compares NameBERT with EthnicSeer (12 classes) and NamePrism (39 classes) under matched label mappings. Under both reduced taxonomies, NameBERT achieves substantially higher accuracy and Macro-F1 than the corresponding baselines, indicating that its performance advantage persists under coarser label spaces.}

Overall, the NameBERT models outperform SOTA baselines on the OAG test set. This superior performance will be validated further for out-of-domain data as well (Section~\ref{sec:out_of_sample}).


\subsubsection{Effect of Label Validation and Data Augmentation}

\textbf{Validated real names (\texttt{test\_filter}).} 
\mc{As shown in Table~\ref{tab:combined_results}, evaluating on \texttt{test\_filter} changes the metric profile of NameBERT$_{aug}$ relative to its performance on \texttt{test\_oag} (Table~\ref{tab:combined_results_oag}):
Accuracy and Weighted-F1 decrease while Macro-F1 increases. 
This behavior is consistent with \texttt{test\_filter} being a label-screened, more balanced per-country evaluation split (retaining up to 1{,}000 names per country), where head classes are less dominant and metrics are more sensitive to minority-country performance. Under this setting, NameBERT$_{aug}$ shows a modest but consistent improvement over NameBERT$_{oag}$ in Macro-F1 (except in the 12-class setting, where head and tail countries are largely mixed), and both NameBERT variants substantially outperform the baselines across all metrics.}

\begin{table*}[htbp] 
\footnotesize
\centering
\setlength{\tabcolsep}{5pt} 
\vspace{-0.2cm}
\begin{tabular}{l c c c c c c c}
\toprule
 & & \multicolumn{3}{c}{\texttt{test\_filter}} & \multicolumn{3}{c}{\texttt{test\_filter\_aug}} \\
\cmidrule(lr){3-5} \cmidrule(lr){6-8}
\textbf{Model} & \textbf{Classes} & \textbf{Acc} & \textbf{W-F1} & \textbf{M-F1} & \textbf{Acc} & \textbf{W-F1} & \textbf{M-F1} \\
\midrule
EthnicSeer & 12 & 0.648 & 0.647 & 0.568 & 0.709 & 0.709 & 0.589\\
NameBERT$_{oag}$ & 12 & \textbf{0.824} & \textbf{0.840} & \textbf{0.796} & 0.735 &0.772 & 0.746\\
{NameBERT$_{aug}$} & 12 & 0.821 & 0.839 &\textbf{0.796} & \textbf{0.849} & \textbf{0.873} & \textbf{0.820}\\
\midrule
NamePrism & 39 & 0.590 & 0.590 & 0.491 & 0.483 & 0.484 & 0.487\\
NameBERT$_{oag}$ & 39 & \textbf{0.802} & 0.800 & 0.712 & 0.698 & 0.685 & 0.633\\
NameBERT$_{aug}$ & 39 & 0.801 & \textbf{0.802} & \textbf{0.723} & \textbf{0.865} & \textbf{0.867} & \textbf{0.815}\\
\midrule
Name2Nat & 99 & 0.325 & 0.306 & 0.163 & 0.218 & 0.187 & 0.141\\
NameBERT$_{oag}$ & 99 & 0.651 & 0.630 & 0.416 & 0.466 & 0.417 & 0.350 \\
{NameBERT$_{aug}$} & 99 & \textbf{0.652} & \textbf{0.635} & \textbf{0.442} & \textbf{0.742} & \textbf{0.732} & \textbf{0.570} \\
\bottomrule
\end{tabular}
\vspace{-0.2cm}
\caption{Performance on \texttt{test\_filter} (LLM-screened real names) and \texttt{test\_filter\_aug} (synthetic-mixed tail diagnostic) under the 99-class taxonomy.  W=Weighted, M=Macro.}
\vspace{-0.3cm}
\label{tab:combined_results}
\end{table*}

\textbf{Validated + augmented tail set (\texttt{test\_filter\_aug}).} 
\mc{Table~\ref{tab:combined_results} shows a sharper model-dependent effect on \texttt{test\_filter\_aug}, which mixes validated real names with synthetic tail-country names. This split deliberately increases tail-class support to quantify long-tail behavior under controlled conditions, which allows stable measurement of tail Macro-F1 and error collapse patterns. Both NameBERT models perform much better than the baselines. Between them, NameBERT$_{oag}$ is worse,
indicating a distribution mismatch under tail-heavy evaluation. In contrast, augmentation yields large gains for NameBERT$_{aug}$ (Macro-F1 0.350$\rightarrow$0.570).}
\mc{As evaluation becomes more balanced or tail-emphasized, accuracy-style metrics may decrease while Macro-F1 better reflects minority-country performance. Augmentation provides a modest tail lift on the real-name split and much larger gains on the synthetic-mixed diagnostic, indicating distribution-dependent benefits.}


\begin{table}[htbp]
\centering
\small
\begin{tabular}{llcc|cc|cc}
\toprule
Split & Model & Acc & M-F1 & Head Acc & Head M-F1 & Tail Acc & Tail M-F1 \\
\midrule
test\_filter & NameBERT$_{oag}$ & 0.651 & 0.416 & \textbf{0.699} & \textbf{0.336} & 0.482 & 0.199 \\
test\_filter & NameBERT$_{aug}$ & \textbf{0.653} & \textbf{0.442} & 0.695 & 0.308 & \textbf{0.505} & \textbf{0.232} \\
\midrule
test\_filter\_aug & NameBERT$_{oag}$ & 0.467 & 0.350 & 0.699 & \textbf{0.336} & 0.282 & 0.158 \\
test\_filter\_aug & NameBERT$_{aug}$ & \textbf{0.742} & \textbf{0.570} & \textbf{0.705} & 0.312 & \textbf{0.773} & \textbf{0.345} \\
\bottomrule
\vspace{-0.6cm}
\end{tabular}
\caption{\mc{Head--tail bucket analysis (tail: countries with <6000 OAG names). Augmentation mainly improves tail performance on the oag-filtered split, while gains on the oag-synthetic-mixed split reflect distribution alignment. M=Macro.}}
\vspace{-0.3cm}
\label{tab:bucket1}
\end{table}

\mc{To further make sense of the results, using our long-tail definition (countries with $<$6{,}000 OAG names), Table~\ref{tab:bucket1} shows that on \texttt{test\_filter} augmentation leaves overall accuracy of the two NameBERT models nearly unchanged (0.651$\rightarrow$0.653) but NameBERT$_{aug}$ improves tail metrics (tail Acc 0.482$\rightarrow$0.505; tail Macro-F1 0.199$\rightarrow$0.232) while slightly reducing head Macro-F1 (0.336$\rightarrow$0.308), suggesting a head-tail trade-off. On \texttt{test\_filter\_aug}, NameBERT$_{aug}$ gains are much larger (Macro-F1 0.350$\rightarrow$0.570), consistent with distribution alignment when synthetic tail names appear at evaluation time. }

\subsubsection{Evaluation on Out-of-Domain Datasets\label{sec:out_of_sample}}

\mc{So far we have been testing on datasets that are either OAG or OAG-augmented. To test cross-domain generalization ability of NameBERT models, we evaluate all models on datasets that are out-of-domain relative to NameBERT.} \yifan{We note that of the three baseline models, only the training data of Name2Nat (the NaNa dataset) is publicly available, which we use as an out-of-domain test set for NameBERT, EthnicSeer and NamePrism. More broadly, the scarcity of publicly available nationality classification datasets limits our capability to conduct an evaluation that is out-of-domain for all models tested. Therefore we also use the LLM-generated \texttt{test\_gold} as our second dataset that is out-of-domain to all but the NameBERT$_{aug}$, which we exclude in the comparison.}


\begin{table}[htbp]
\centering
\small
\begin{tabular}{lccc ccc ccc}
\toprule
& \multicolumn{3}{c}{\textbf{12 Classes}} 
& \multicolumn{3}{c}{\textbf{39 Classes}} 
& \multicolumn{3}{c}{\textbf{99 Classes}} \\
\cmidrule(lr){2-4} \cmidrule(lr){5-7} \cmidrule(lr){8-10}
\textbf{Model} 
& \textbf{Acc} & \textbf{W-F1} & \textbf{M-F1}
& \textbf{Acc} & \textbf{W-F1} & \textbf{M-F1}
& \textbf{Acc} & \textbf{W-F1} & \textbf{M-F1} \\
\midrule

EthnicSeer
& 0.748 & 0.760 & 0.622
& -- & -- & --
& -- & -- & -- \\

NamePrism
& -- & -- & --
& 0.609 & 0.647 & 0.113
& -- & -- & -- \\

Name2Nat
& -- & -- & --
& -- & -- & --
& \textbf{0.958}${}^*$ & \textbf{0.971}${}^*$ & \textbf{0.582}${}^*$ \\

NameBERT$_{oag}$
& \textbf{0.807} & \textbf{0.831} & \textbf{0.734}
& \textbf{0.783} & \textbf{0.792} & \textbf{0.503}
& 0.552 & 0.540 & 0.338 \\

NameBERT$_{aug}$
& 0.799 & 0.826 & 0.730
& 0.777 & 0.789 & 0.465
& 0.544 & 0.539 & 0.300 \\

\bottomrule
\end{tabular}
\vspace{-0.4cm}
\caption{Performance comparison on the \texttt{NaNa\_test}. ${}^*$\emph{Name2Nat was trained on the full NaNa dataset}; its results are therefore not out-of-sample in this evaluation.}
\label{tab:nana_test}
\end{table}

\mc{\textbf{Wikipedia names (\texttt{NaNa\_test}).} The NaNa dataset~\cite{park2020name2nat} consists of real-world names collected from Wikipedia. 
This dataset is out-of-domain for NameBERT, EthnicSeer and NamePrism, but not for Name2Nat. As shown in Table~\ref{tab:nana_test}, both NameBERT models substantially outperform EthnicSeer (12 classes) and NamePrism (39 classes), indicating that representations learned from the OAG large-scale author metadata transfer to Wikipedia names when the label space is moderately coarsened. At 99 classes, performance of NameBERT is reduced, reflecting the intrinsic difficulty of fine-grained nationality prediction under domain shift. 
We mark Name2Nat results with an asterisk because Name2Nat was trained on the full NaNa dataset, therefore is not measured out-of-domain in this setting.} 

\mc{Among the NameBERT models, NameBERT$_{oag}$ is slightly better. To understand why augmentation does not help the overall performance, we performed a head-tail test in Table~\ref{tab:namebert_head_tail}. It is seen that NameBERT$_{aug}$ does improve tail accuracy (0.414$\rightarrow$0.429) and macro-F1 (0.166$\rightarrow$0.167), but overall accuracy and Macro-F1 decrease (0.552$\rightarrow$0.544; 0.338$\rightarrow$0.300), suggesting that cross-domain benefit from synthetic augmentation is limited to tail countries, likely because the augmented data shifts the underlying country distribution.}

\begin{table}[htbp]
\centering
\begin{tabular}{lcc|cc|cc}
\toprule
Model & Acc & M-F1 & Head Acc & Head M-F1 & Tail Acc & Tail M-F1 \\
\midrule
NameBERT$_{oag}$ & \textbf{0.552} & \textbf{0.338} & \textbf{0.570} & \textbf{0.210} & 0.414 & 0.166 \\
NameBERT$_{aug}$ & 0.544 & 0.300 & 0.559 & 0.177 & \textbf{0.429} & \textbf{0.167} \\
\bottomrule
\vspace{-0.6cm}
\end{tabular}
\caption{\mc{NameBERT models on head and tail subsets of the NaNa dataset.}}
\vspace{-0.4cm}
\label{tab:namebert_head_tail}
\end{table}

\begin{table}[htbp]
\centering
\small
\vspace{-0.2cm}
\begin{tabular}{lccc ccc ccc}
\toprule
& \multicolumn{3}{c}{\textbf{12 Classes}} 
& \multicolumn{3}{c}{\textbf{39 Classes}} 
& \multicolumn{3}{c}{\textbf{99 Classes}} \\
\cmidrule(lr){2-4} \cmidrule(lr){5-7} \cmidrule(lr){8-10}
\textbf{Model} 
& \textbf{Acc} & \textbf{W-F1} & \textbf{M-F1}
& \textbf{Acc} & \textbf{W-F1} & \textbf{M-F1}
& \textbf{Acc} & \textbf{W-F1} & \textbf{M-F1} \\
\midrule
EthnicSeer 
& 0.763 & 0.778 & 0.628 
& -- & -- & -- 
& -- & -- & -- \\

NamePrism 
& -- & -- & -- 
& 0.680 & 0.670 & 0.617 
& -- & -- & -- \\

Name2Nat 
& -- & -- & -- 
& -- & -- & -- 
& 0.238 & 0.186 & 0.177 \\

NameBERT$_{oag}$ 
& \textbf{0.785} & \textbf{0.812} & \textbf{0.722}
& \textbf{0.722} & \textbf{0.712} & \textbf{0.641}
& \textbf{0.402} & \textbf{0.352} & \textbf{0.352} \\

\bottomrule
\end{tabular}
\vspace{-0.2cm}
\caption{Synthetic stress test on \texttt{test\_gold}.}
\label{tab:synthetic_stress}
\end{table}
\vspace{-0.2cm}

\textbf{Synthetic names (\texttt{test\_gold}).} \mc{We further evaluate models on \texttt{test\_gold}, which contains 100 LLM-generated names per country and is intended as a controlled stress test rather than a proxy for real-world deployment.
To avoid confounding from stylistic overlap between synthetic training and synthetic testing, we report only NameBERT$_{oag}$ and other baselines on \texttt{test\_gold}, and this split is considered out-of-domain for all of them.
Table~\ref{tab:synthetic_stress} summarizes results across three taxonomies. NameBERT$_{oag}$ consistently outperforms the baselines, with the largest gaps at finer granularities (39 and 99 classes), suggesting stronger name representations under this controlled stress-test setting.}

\mc{In summary, NameBERT models perform significantly better than the three state-of-the-art baselines when all models are compared out-of-domain. In the same setting, LLM-based augmentation provides benefit to the tail countries, at the expense of overall performance.}

\subsubsection{Throughput and Latency Comparison with LLMs}


To evaluate computational efficiency, we benchmark NameBERT against several LLM APIs, including GPT-5, Claude Sonnet 4.5, and Gemini 3. Experiments for NameBERT were run locally on a consumer-grade NVIDIA RTX 3070 Laptop GPU (8.6\,GB VRAM), while LLMs were accessed via their cloud APIs. Table~\ref{tab:benchmark_short} summarizes throughput and per-name latency across batch sizes. {Latency is reported as average batch runtime divided by batch size. LLM costs are based on API list pricing at the time of experiments.}


LLM throughput ranges from 3.7 to 4.8 names/s.
In contrast, NameBERT achieves 1{,}831 names/s at batch size 1{,}000 and sustains 1{,}419 names/s at batch size 10{,}000, corresponding to roughly two orders of magnitude higher throughput than LLM APIs in our measurements. Per-name latency for NameBERT drops below 1\,ms for batch sizes of 100 or more, compared to 200--270\,ms/name for LLM APIs at batch size 1{,}000.

Overall, these results show that a local, specialized classifier offers substantial throughput and cost advantages over LLM-based alternatives when processing millions of names at scale, and also delivers superior performance in repeated use due to its lower latency.

\begin{table}[t]
\vspace{-0.4cm}
\setlength{\tabcolsep}{2pt}

\centering
\small
\begin{tabular}{lccccc}
\toprule
\textbf{Model} & \textbf{Type} & \textbf{Batch} & \textbf{Throughput (names/s)} & \textbf{Latency (ms/name)} & \textbf{\$/1M names} \\
\midrule
Sonnet 4.5 & LLM API & 1{,}000 & 4.3 & 234.6 & 282 \\
Gemini 3 & LLM API & 1{,}000 & \textbf{4.8} & 208.7 & 824 \\
GPT-5 & LLM API & 1{,}000 & 3.7 & 268.4 & 469 \\
\midrule
NameBERT & Local GPU & 1{,}000 & \textbf{1{,}831} & \textbf{0.5} & $\approx 0$ \\
\bottomrule
\vspace{-0.4cm}
\end{tabular}
\caption{Computational efficiency at batch size 1{,}000.}
\vspace{-0.5cm}
\label{tab:benchmark_short}
\end{table}

\subsubsection{Discussions}

\mc{Across increasingly controlled evaluation settings, we observe a consistent pattern: improving label quality and increasing tail coverage through LLM augmentation shifts evaluation emphasis away from head-dominated accuracy toward fairness-oriented metrics such as Macro-F1. 
NameBERT$_{aug}$ shows clearer gains on long-tail metrics in our in-domain validated setting, particularly in tail-country Macro-F1. At the same time, our out-of-domain results on Wikipedia names (NaNa) indicate that tail country gains of NameBERT$_{aug}$ are not as significant, suggesting limited cross-domain generalization from synthetic name generation.} \mc{Taken together, these results indicate that augmentation alone is not a guaranteed solution to long-tail nationality prediction. Instead, its impact is distribution-dependent and interacts with model capacity: NameBERT trained on OAG provides strong performance under both the full 99-class setting and coarser taxonomies, and can benefit from targeted augmentation when evaluation includes tail-heavy name distributions (e.g., synthetic-mixed diagnostics) or when tail classes are under-supported in-domain. A smaller LSTM model (not reported) did not benefit from the augmentation at all.} 

\mc{These findings provide an affirmative answer to \textbf{RQ1}: OAG offers a large and diverse source of proxy nationality labels. For \textbf{RQ2}, we show that NameBERT trained on OAG achieves superior performance compared with state-of-the-art baselines on multiple benchmarks, including out-of-domain ones.
For \textbf{RQ3},
while LLM-based augmentation yields conditional improvements, notably on tail-country metrics in-domain and under tail-aware diagnostic evaluations, it does not consistently improve out-of-domain generalization.}


\begin{figure}[htbp]
  \centering
  \begin{subfigure}[b]{0.48\textwidth}
    \centering
    \includegraphics[width=\textwidth]{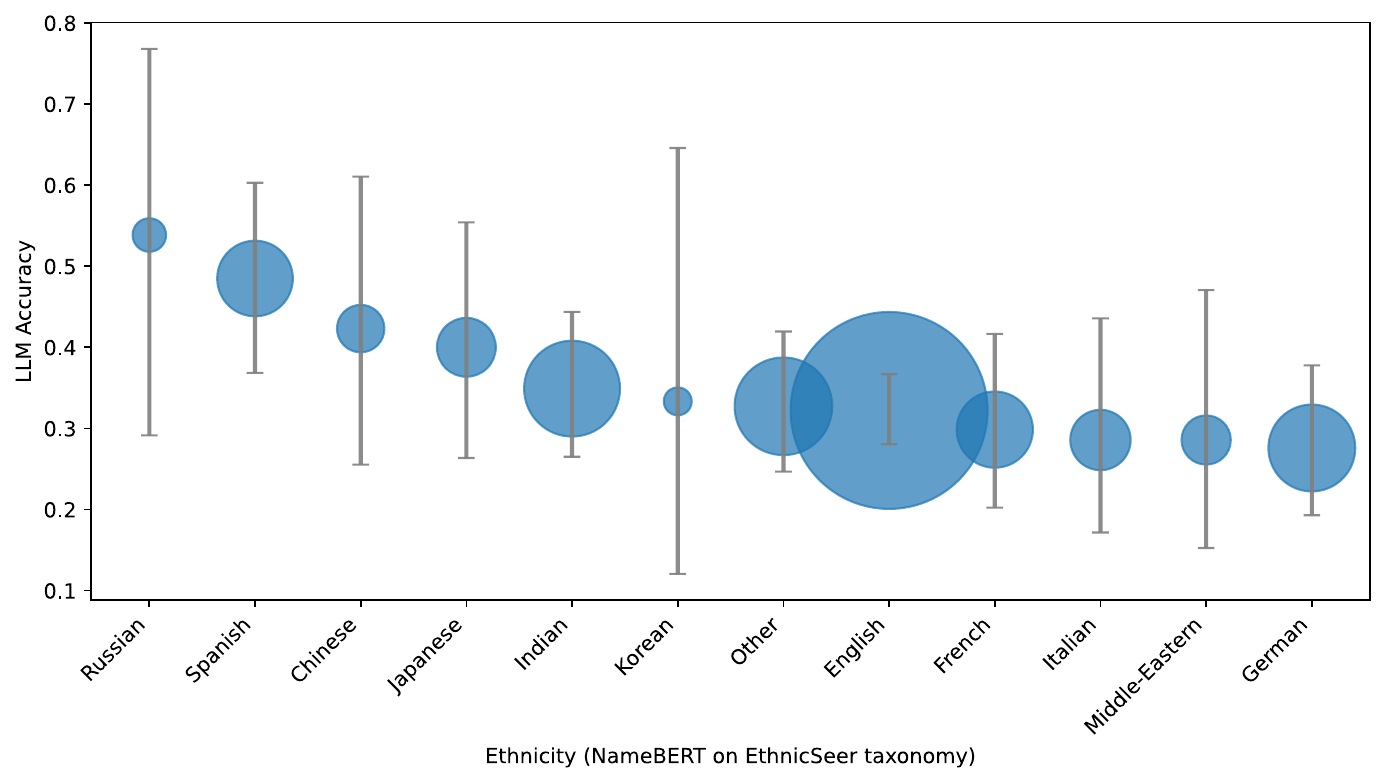}
    \caption{LLM Factual Accuracy by Nationality}
    \label{fig:acc}
  \end{subfigure}
  \hfill
  \begin{subfigure}[b]{0.48\textwidth}
    \centering
    \includegraphics[width=\textwidth]{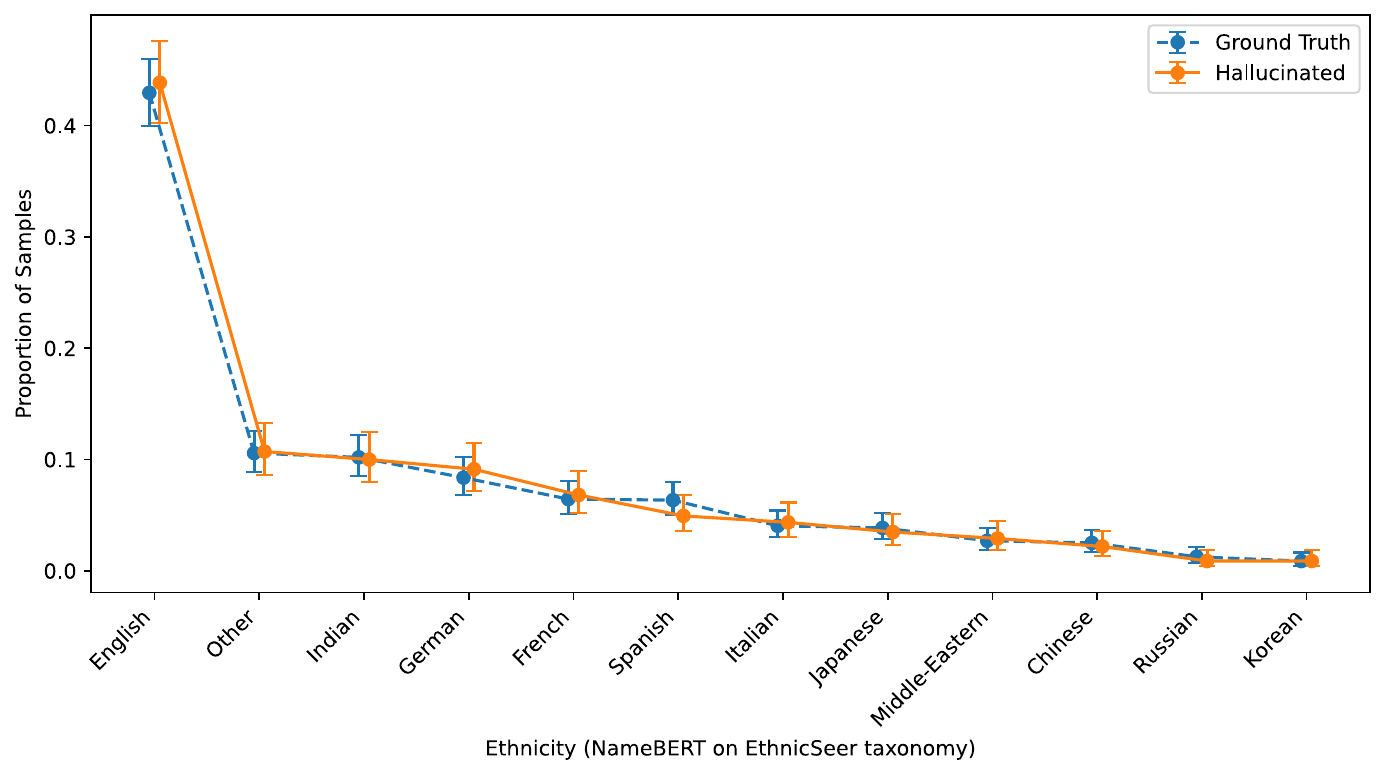}
    \caption{Ethnicity Distribution}
    \label{fig:distribution}
  \end{subfigure}
  \caption{Results of GPT-4o on SimpleQA with NameBERT$_{aug}$ (Wilson 95\% CI)}
  \label{fig:SimpleQA}
\end{figure}
\section{Application Study: Bias Analysis on SimpleQA Name-Based Questions}

Bias analysis is an application of nationality classification. \textbf{SimpleQA}~\cite{wei2024measuring} is a benchmark for evaluating LLM factual recall across 4{,}326 questions. We extract the 1{,}041 questions whose ground-truth answers are human names to investigate whether nationality biases emerge during hallucination, e.g., substitute an English name when the LLM fails to recall a Japanese one. We evaluate GPT-4o, GPT-4.1, and Claude-sonnet-3.7. Claude-sonnet-3.7 refuses $\sim$60\% of questions, so we focus on GPT-4o, which achieves 33.6\% accuracy on our subset (vs.\ 38–40\% on the full benchmark), confirming that name-based questions are systematically harder. We map both ground-truth and hallucinated names to the 12-category EthnicSeer taxonomy using NameBERT$_{aug}$.

Figure~\ref{fig:acc} shows substantial accuracy variation across groups, with the largest (English) group not achieving the highest accuracy, indicating \emph{group frequency alone does not explain performance}. Figure~\ref{fig:distribution} shows that hallucinated-answer distributions largely mirror ground-truth distributions, suggesting GPT-4o's errors \emph{do not induce a strong demographic shift}. Overall, demographic concerns appear to stem from unequal group accuracy rather than skewed hallucination demographics.

\section{Conclusion}
In this work, we construct a large-scale OAG-derived name–nationality dataset (1.34M name–country pairs spanning 99 countries) and develop a transformer-based nationality classifier, NameBERT, that substantially outperforms SOTA baselines on both in-domain and out-of-domain evaluations. We empirically characterize the gains from LLM-based augmentation for tail-country classification -- strong in-domain and under synthetic settings, while still beneficial for out-of-domain tail countries. Compared with LLM API–based classification, locally deployed NameBERT provides orders-of-magnitude higher throughput, lower latency, and no per-query costs, enabling efficient large-scale deployment.
Future work includes aligning classifiers' predictions more closely with LLM-derived nationality judgments while retaining the efficiency advantages of a local model, as well as the application of NameBERT to downstream fairness auditing tasks beyond SimpleQA.




\printbibliography[heading=subbibintoc]

\end{document}